\documentclass{article}

\PassOptionsToPackage{sectionbib,numbers,compress}{natbib}

\usepackage[preprint]{nips_2018}

\usepackage[utf8]{inputenc} 
\usepackage[T1]{fontenc}    
\usepackage{hyperref}       
\usepackage{url}            
\usepackage{booktabs}       
\usepackage{amsfonts}       
\usepackage{nicefrac}       
\usepackage{microtype}      
\usepackage{amsmath}
\usepackage[pdftex]{graphicx}
\usepackage{ulem}
\usepackage{wrapfig}

\title{Reliable counting of weakly labeled concepts by a single spiking neuron model.}

\author{
  Hannes~Rapp \\
  Computational Systems Neuroscience \\
  University of Cologne\\
  Cologne, Germany \\
  \texttt{hannes.rapp@smail.uni-koeln.de} \\
  \And
  Martin~Paul~Nawrot \\
  Computational Systems Neuroscience \\
  University of Cologne \\
  Cologne, Germany \\
  \texttt{mnawrot@uni-koeln.de} \\
  \And
  Merav~Stern \\
  Department of Applied Mathematics \\
  University of Washington \\
  Seattle, WA 98195 \\
  \texttt{ms4325@uw.edu} \\
}

\begin{document}

\maketitle

\begin{abstract}
  Making an informed, correct and quick decision can be life-saving. It's crucial for animals during an escape behaviour or for autonomous cars during driving. The decision can be complex and may involve an assessment of the amount of threats present and the nature of each threat. Thus, we should expect early sensory processing to supply classification information fast and accurately, even before relying the information to higher brain areas or more complex system components downstream.
 Today, advanced convolutional artificial neural networks can successfully solve visual detection and classification tasks and are commonly used to build complex decision making systems. However, in order to perform well on these tasks they require increasingly complex, "very deep" model structure, which is costly in inference run-time, energy consumption and number of training samples, only trainable on cloud-computing clusters.
  A single spiking neuron has been shown to be able to solve recognition tasks for homogeneous Poisson input statistics, a commonly used model for spiking activity in the neocortex, including the visual cortex, when modeled as leaky integrate and fire with gradient decent learning algorithm it was shown to posses a variety of complex computational capabilities. Here we improve its implementation. We also account for more natural stimulus generated inputs that deviate from this homogeneous Poisson spiking. The improved gradient-based local learning rule allows for significantly better and stable generalization and more efficient performance. We also show that with its improved capabilities it can count weakly labeled visual concepts by applying our model to a problem of multiple instance learning (MIL) with counting where labels are only available for collections of concepts. In this visual counting MNIST task the neuron exploits the improved implementation and outperforms conventional ConvNet architecture with similar parameter space size and number of training epochs.
\end{abstract}
\clearpage

\section{Introduction}
   Visual nervous systems are highly efficient in solving complex classification and detection tasks. During evolution, they have grown to solve problems requiring fast decoding while being precise and energy efficient. Remarkably, different visual systems of different animal species found different solutions to these requirements, specialized to their own natural habitat. Some of the properties of the capabilities of these visual systems are very appealing for programming applications like robotics and autonomous driving. However, these are not yet implemented efficiently in our computers, compared to living systems. Therefore it is worth to explore fundamentally different computing models that exploit biological mechanisms of visual information processing. 
   
   The basic elements of the visual system are neurons. Biological neurons communicate among themselves with discrete time events - the so-called spikes. However, networks of spiking neurons are difficult to model and analyze, because of the discrete nature of spikes and their mechanism of fast rise and reset of the neuron's voltage. Hence, vast majority of models ignore the spikes discrete nature and assume that only the rate of spike occurrences matters. Rates, by concept, can be treated with continuous time-varying functions, which allows for various derivative based approaches such as gradient decent learning, to be implemented. 
   
   Hence, it is not surprising that the majority of neural network studies and algorithms are rate based. Their implementations through deep learning (\cite{lecun2015deep}, \cite{schmidhuber2015deep}), ConvNet (\cite{lecun1998gradient}), echo state (\cite{Jaeger:2004}) and recurrent Long Short-Term Memory (LSTM) networks (\cite{Hochreiter:1997}) are indeed highly successful. As the tasks are becoming more complex, however, these model classes are becoming increasingly more costly and often require cloud-computing clusters and millions of samples to be trained \cite{NIPS2012_4824}, \cite{Simonyan:2014rt}. It was recently shown \cite{OpenAI:compute} that the amount of computation needed by such artificial systems has been growing exponentially since 2012. 
   
   Therefore, it is worth examining biologically realistic spiking neurons as computational units despite technical challenges. Thanks to their efficiency, spiking neurons and networks are natural candidates for the next generation of neural network algorithms. Some recent studies managed to train spiking neural networks with gradient-based learning methods. To overcome the discontinuity problem, the currents  created by the spikes in receiving neurons (essentially through linear low-pass filtering) were used in \cite{Clopath:2017} and \cite{Huh:2017dz} for the training procedures. Other studies use the timing of the spikes as a continues parameter \cite{Bohte2000}, \cite{OConnor:2017xd}, which leads to neuronal (synaptic) learning  rules that rely on the exact time intervals between the spikes of the sending and receiving neurons (pre- and post- synaptic). These Spike Timing Dependent Plasticity (STDP) rules had first been observed experimentally and hence much of attention is given to them in neuroscientific studies \cite{Bi:2001} \cite{Caporale:2008} \cite{SongSTDP:2000}. But their computational capability, especially for classification tasks, has not been well exploited. It is generally interesting, still highly debated question, whether the brain uses the timing of the spikes or their rate to represent the information, and whether connectivity is modified in the brain accordingly. We leave this broad open question, outside the scope of this paper.
   
   An additional intriguing approach, is to train spiking neurons as classifiers, perceptron-like machines \cite{Gutig:2006lf}, \cite{Memmesheimer:2014}. Here, the gradient learning is done based on the neuron's membrane voltage in relation to the maximum voltage the neuron reached, compared to its threshold for spiking. A full spiking network was trained in a similar fashion to generate patterns \cite{SussilloCOSYNE:2018}. Here we concentrate on the algorithm for the recently published Multi-Spike Tempotron \cite{Gutig:2016}, a single neuron leaky integrate and fire model that solves regression problems including learning how to recognize concepts within a collection. Specifically, the Multi-Spike Tempotron (MST) learns to generate a certain number of spikes for a given concept (stimulus). The learning algorithm changes the input weights according to a voltage threshold gradient decent, such that the weights eventually fit the threshold in which the neuron generates the exact number of spikes required. We outline the underlying algorithm in more detail in the method section. The signals we use for training the Multi-Spike Tempotron are for collections (bags) of concepts, a learning strategy termed \textit{Multiple Instance Learning (MIL) with counting} that has been recently proposed in the literature \cite{Foulds:2010}, \cite{Lempitsky2010}, \cite{Segui:2015}. Thus, the Multi-Spike Tempotron is capable of evaluating a sum of multiple object instances present in an input stream. This is especially useful in early stages of decision making \cite{Stanford:2010}, where assessment of the number of threats present is needed quickly, for example to help escape predators or avoid collisions.  
   
   It has been shown that in-vivo cortical spiking activity is typically more regular then Poisson \cite{Mochizuki:2016}, \cite{Nawrot:2010}. In general any correlated stimuli input is expected to deviate from Poisson \cite{Farkhooi:2011}. Moreover, input is generally non-homogenous, i.e. time-varying. However only homogeneous Poisson statistic of input patterns and background were considered in \cite{Gutig:2016}. Hence, here we study learning capabilities of the Multi-Spike Tempotron when the statistic of input patterns deviates from the Poisson statistic and when the background statistic is non-homogeneous, i.e. time-varying as under realistic noisy conditions.
   
   Our main goal in this study is to improve the capabilities of the Tempotron by improving its implementation. The Tempotron algorithm originally used Momentum \cite{POLYAK:1964} to boost its capabilities. We show here that a connection (synapse) specific adaptive update approach with smoothing over previous updates, similar to RMSprop \cite{TielemanHinton:2012}, generates significantly better and stable generalization and more efficient performance of the learning capabilities of the Multi-Spike Tempotron. We review both the Momentum and the RMSprop in the method section. We further show that our improved learning algorithm performs better in the biological context of non-homogeneous spiking in the visual cortex as well as in a counting task on MNIST figures. We finally show that it outperforms a deep network (ConvNet) with similar parameter space and training epochs.  
\section{Method}

\subsection{Tempotron Model}
The Multi-Spike Tempotron is a current-based leaky integrate-and-fire neuron model. Its membrane potential, $V(t)$, follows the dynamical equation:
\begin{align}
  V(t) = V_{rest} + \sum^{N}_{i=1} \omega_i \sum_{t^{j}_{i} < t} K(t-t^{j}_{i}) - \vartheta \sum_{t^{j}_{spike} < t} e^{-\frac{t-t^{j}_{spike}}{\tau_m}}
\end{align}
where $t^{j}_{i}$ denotes the time of spike number $j$ from the input source (presynaptic) number $i$, and $t^j_{spike}$ denotes the time of spike number $j$ of the Tempotron neuron model. Every input spike at $t^{j}_{i}$ contributes to the potential by the kernel:    
\begin{align}
  K(t-t^{j}_{i}) &= V_{norm} (e^{- \frac{t-t^{j}_{i}}{\tau_m}} - e^{-\frac{t-t^{j}_{i}}{\tau_s}})
\end{align}
times the synaptic weight of that input source $\omega_i$. These synaptic input weights are learned via the gradient decent algorithm. The kernel is normalized to have its peak value at $1$ with $V_{norm} = \eta^{\eta/(\eta-1)}/(\eta-1)$ and $\eta = \tau_m/\tau_s$ where $\tau_m$ and $\tau_s$ are the membrane time constant and the synaptic decay time constant. The kernel is causal, it vanishes for $t<t_i^j$. When $V(t)$ crosses the threshold $\vartheta$ the neuron emits a spike and is reset to $V_{rest} = 0$ by the second term in equation (1).

In order to have the neuron emit the required number of spikes in response to a specific concept (implemented as synaptic input spike pattern) the weights $\omega_i$ are modified. Since the required spike numbers are non differentiable discrete numbers the gradient for the weights is derived from the spiking threshold. We wish to change the weights such that the neuron's voltage would reach a critical threshold $\vartheta^{*}_{k}$ that would coincide with its threshold $\vartheta$, that would be crossed exactly $k$ times to generate the $k$ desired spikes. This loss function is called Spike-Threshold Surface (STS). Hence the appropriate gradient can be describe by:
\begin{align}
  \Delta \omega = \eta \lambda \vec{\nabla}_{\vec{\omega}} \vartheta^{*}_{k}
\end{align}
Where $\eta \in \{-1,1\}$ controls whether to increase or decrease the number of output spikes towards the $k$ required, $\lambda$ is the learning rate parameter that controls the size of the gradient step and $\vec{\nabla}_{\vec{\omega}} \vartheta^{*}_{k}$ is the gradient of the critical voltage threshold with respect to the synaptic weights. In practice, multiple concepts are presented and the learning signal is the sum of spikes that are required by the collection of concepts to generate. 

To evaluate the expression in (3) we use the properties of the $k$-th spike time $t^{*}_{k}$ in which the potential reaches the critical threshold $\vartheta^{*}_{k}$ and hence $\vartheta^{*}_{k} = V(t^{*}_{k}) = V(t^{j}_{s})$ where $t^{j}_{s} < t^{*}_{k}$ are all the previous time points when the neuron spiked. Together with the voltage (membrane potential) dynamics (1) and (2) a recursive expression, that depends on all previous spike times, can be found for the gradient (3). For details about the full derivation see the $\vartheta^{*} gradient$ section in the methods of \cite{Gutig:2016}. 

The learning rate $\lambda$ is global for all synaptic weights. Hence, the gradient descent takes an equal size step along all directions. If this parameter is too small the training process will take very long, but if it's too big the algorithm might miss an optima within the error surface and never converge to a desired solution. Hence, tuning this learning rate is important to achieve decent training speed.

A possible approach \cite{Gutig:2016} to avoid these problems is to update the weights according to the accumulating error, \textit{Momentum} heuristic:  
\begin{align}
\Delta \omega &= \alpha \Delta \omega(t-1) + \Delta \omega(t) \nonumber  \\
                &= \alpha \Delta \omega(t-1) + \eta \lambda \vec{\nabla}_{\vec{\omega}} \vartheta^{*}_{k}, 
\end{align} 
where $\alpha$ is the \textit{Momentum} parameter. 

\subsection{Adaptive input weight learning and gradient smoothing}
We propose here to use an adaptive learning approach for the weight updates. The algorithm fits each input synapse with its own update rate and by doing so it takes into account that each synapse contributes to the overall update with a different level of importance. For example, updates should be larger for directions which provide more consistent information across examples.
The RMSprop  (Root Mean Square (back-)propagation) \cite{TielemanHinton:2012} is a possible approach to achieve this. It was successfully used in deep learning for training mini-batches. It computes an adaptive learning rate per synapse weight $\omega_i$ as a function of its previous gradient steps :
\begin{align}
v_{i}(t) &= \gamma v_{i}(t-1) + (1-\gamma) (\Delta \omega_{i}(t))^2 \nonumber \\
\Delta \omega_{i}(t) &= \frac{\eta \lambda}{\sqrt{v_{i}(t)}} \nabla_{\omega_i} \vartheta^{*}_{k} 
\end{align} 

\section{Results}
We evaluate the learning performance of the Multi-Spike Tempotron with Momentum and adaptive learning (RMSprop) on the biologically relevant problem of generic, task-related spiking activity-like in the visual cortex (in particular, as part of the neocortex in general) and afterwards turn to the applied problem of visual counting of digits.
We first consider the biological application and evaluate the model under different input statistics that deviate from homogeneous Poisson. We do this by constructing data-sets where the task-related patterns to be learned are drawn from three different distributions. We then slowly increase the complexity of noise level by jittering the spikes or by adding a background activity which is either stationary or time-varying. For this we construct several synthetic data-sets with varying noise levels.
We then apply the MST model to a visual detection problem of counting even digits within an image composed of several random handwritten MNIST digits and compare performance with a conventional Convolutional Neural Networks.
All simulations are carried out with the same set of parameters, $V_{rest}=0, \quad \vartheta=1, \quad \eta=0.001, \tau_{s}=0.005, \quad \tau_{m}=0.015, \quad dt=10^{-3}$sec, using our discrete-time implementation of the Multi-Spike Tempotron in MATLAB. To the best of our knowledge this is also the first publicly available implementation of the Multi-Spike Tempotron model.

\begin{figure}[ht]
  \centering
  \includegraphics[width=0.8\linewidth]{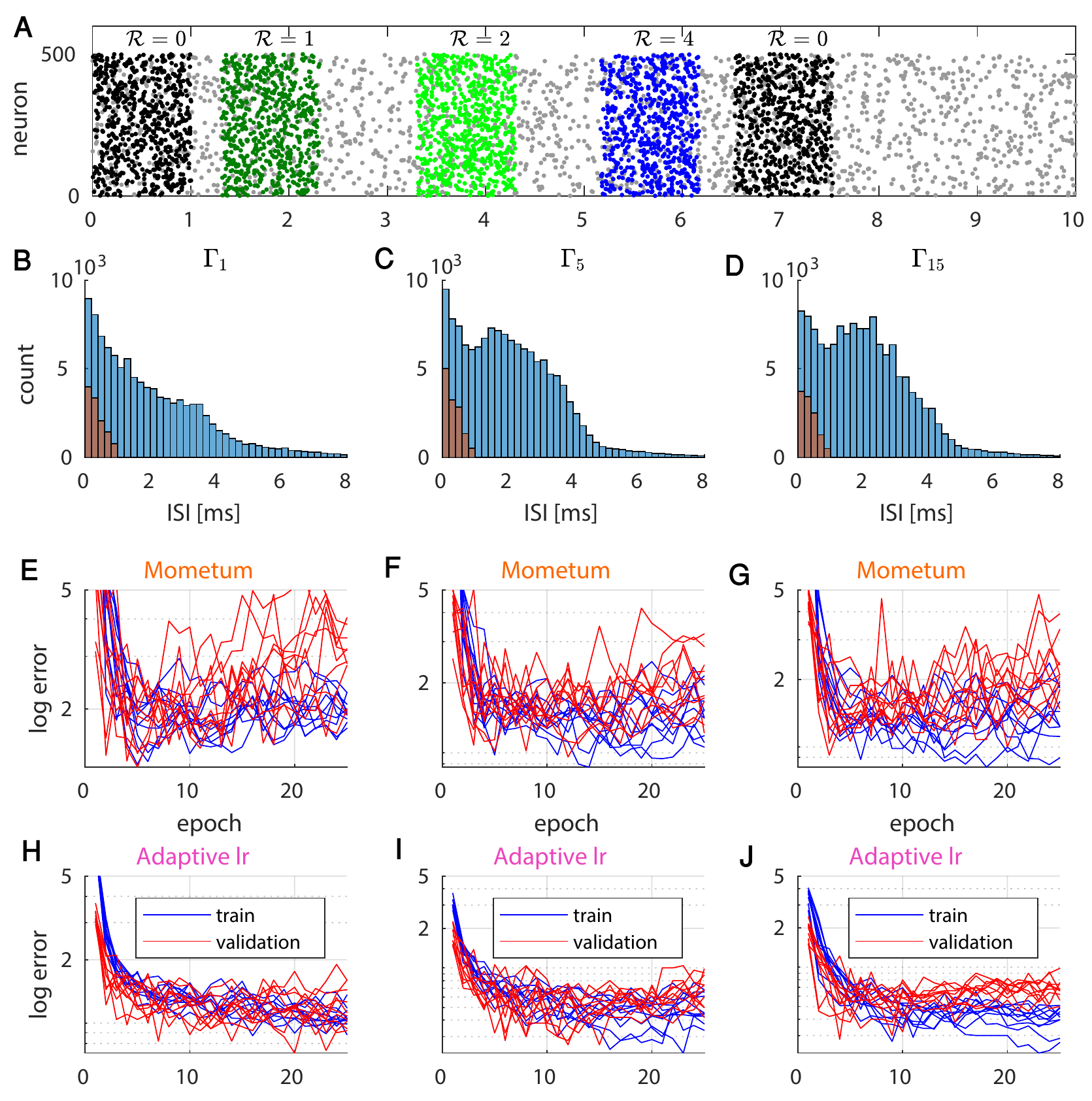}
  \caption{Training convergence for momentum and adaptive learning in multiple background statistics. 
  \textbf{A.} A $10$sec duration spike train input example. The spike train is composed of three patterns, each with a distinct reward (dark green, green, blue), background activity (gray) and two distracting patterns (black). The MST is supposed to fire $\Sigma_i {\mathcal{R}_i} = 7$ spikes over the whole sequence, whereby $\mathcal{R}=0$ spikes for the distracting patterns and $\mathcal{R} \in \{1,2,4\}$ for the colored patterns. 
  \textbf{B-D.} Inter-spike interval (ISI) histograms for three different input statistics data sets. The histograms are shown for the complete input sequence (blue) and patterns only (brown). Patterns are drawn from different processes: \textbf{B.} ($\Gamma_1$ (Poisson), \textbf{C.} $\Gamma_5$, and \textbf{D.} $\Gamma_{15}$). The patterns are embedded in $10$sec inhomogeneous Poisson background activity.
  \textbf{E-J.} Learning curves (blue) and validation curves (red) for the \textbf{E,H.} ($\Gamma_1$ (Poisson), \textbf{F,I.} $\Gamma_5$, and \textbf{G,J.} $\Gamma_{15}$) patterns statistics, with MST \textbf{E-G.} momentum-based learning implementation \cite{Gutig:2016} and \textbf{H-J.} adaptive learning implementation. Each statistic and implementation includes 10 independent simulation results. Learning convergence shows significantly more variance when using momentum compared to adaptive learning. The same is true for the validation error. These suggest that adaptive learning is capable of finding wider optima as compared to momentum.}
  \label{fig:convergence_msp}
\end{figure}

\subsection{Task-related inhomogenous activity in the visual cortex}
\begin{figure}[ht]
  \centering
  \includegraphics[width=0.8\linewidth]{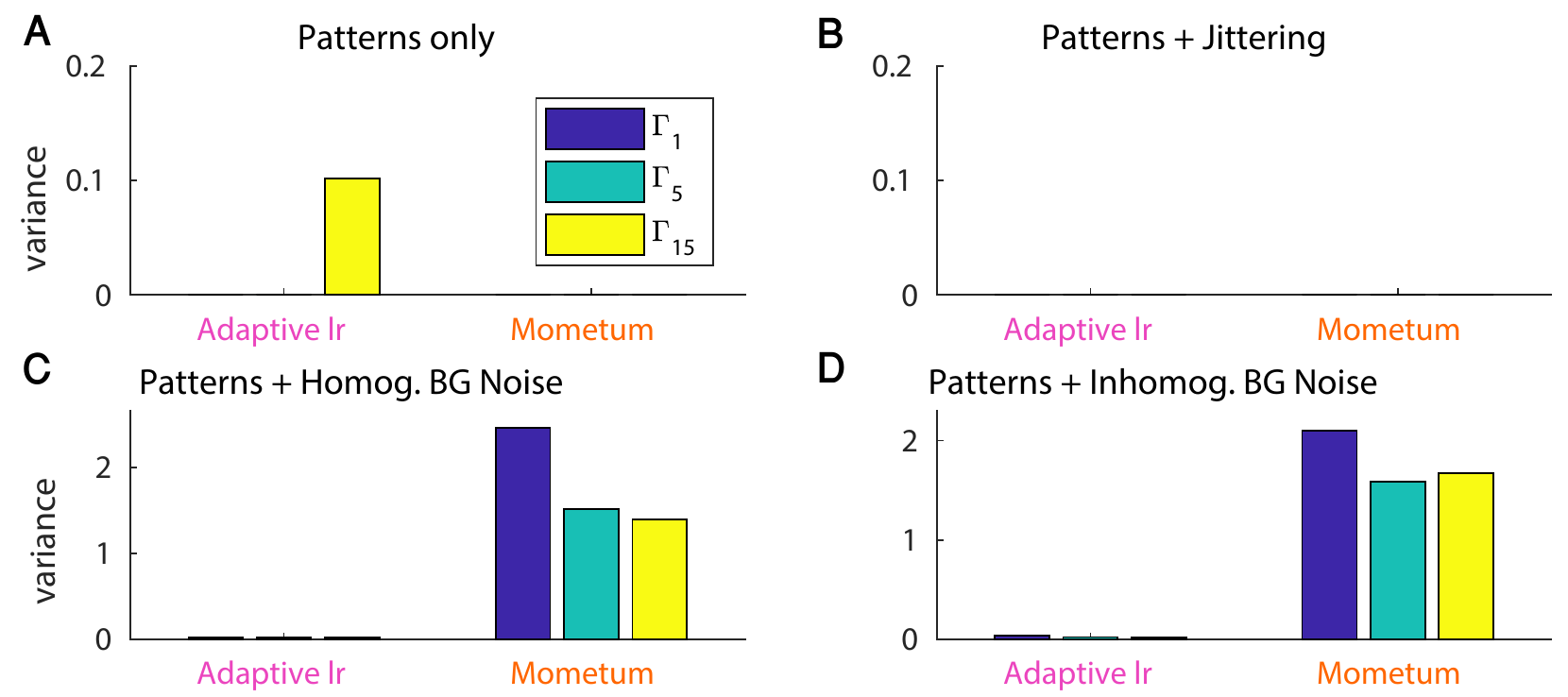}
  \caption{Generalization error variances for different data-sets with increasing noise complexity. For each data set the variance of generalization error is shown for Momentum and adaptive learning (lr). Variance is measured after 10 training epochs where convergence is considered to be reached by both methods.}
  \label{fig:summary_variance}
\end{figure}

We construct three data-sets, each including $9$ generated patterns. Out of this 9 patterns, 5 patterns are considered to be \textit{task-related} and are associated with some positive reward $\mathcal{R}$. The remaining 4 patterns are considered to be distractor patterns with reward $0$. All patterns are generated as 1sec long spike trains by drawing instantaneous firing rates from three different stationary point processes (renewal processes): $\Gamma_1$ representing the homogeneous Poisson process, $\Gamma_5$, and $\Gamma_{15}$ with a fixed intensity (or rate) of $\lambda=0.89$ spike events per second (fig. \ref{fig:convergence_msp} B-D.). Each pattern is associated with a fixed, positive integer reward $\mathcal{R}_{i} \in [0,9]$. Input spike trains of $10$ sec are assembled by drawing a random number of patterns from a Poisson distribution of mean $5$ patterns (with replacement). These patterns are randomly positioned within those $10$ sec but are not allowed to overlap (an example of an input spike-train is shown in fig. \ref{fig:convergence_msp} A). The training target for each of such input spike train is determined as the sum over all individual rewards $\Sigma_i {\mathcal{R}_{i}}$ of each occurring pattern. We evaluate learning under different noise levels (fig. \ref{fig:summary_variance} bottom): \textit{patterns only}, \textit{patterns + spike jittering}, \textit{patterns + homogenous Poisson background activity} and finally \textit{patterns + inhomogenous Poisson background activity}. The homogenous background activity is drawn from a stationary Poisson process while for the inhomogenous case the instantaneous firing rates are modulated by superimposed sinusoidal functions.
While speed of convergence is similar, we find that using adaptive learning results in significantly less variant training error . This also holds for variance of test error on an independent validation data-set and results in better generalization capabilities to previously unseen inputs (\textit{patterns only} fig. \ref{fig:summary_variance}). The adaptive, per synapse learning rate combined with smoothing over past gradients has a regularizing effect and prevents the model from over-fitting to the training data. We further conclude, that the modified algorithm is able to find better and wider optima of the spike-threshold surface loss function as compared to learning with Momentum. More importantly this behaviour is consistent and independent of the input spike train's distribution and noise level (fig. \ref{fig:summary_variance}).

\subsection{Counting MNIST}
\begin{wrapfigure}{r}{0.3\textwidth}
  \begin{center}
  \includegraphics[width=0.3\textwidth]{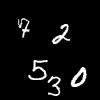}
  \caption{Example from the Counting MNIST data set. The model should learn to count the number of even digits in a given image (in this case 2).}
  \label{fig:MNISTsample}
  \end{center}
\end{wrapfigure}

In this section we consider the applied problem of visual detection, namely the problem of multiple-instance learning using the MNIST \cite{lecun-mnisthandwrittendigit-2010} data-set of handwritten digits. Following  \cite{Segui:2015}, \cite{Fomoro} we generate new images of size 100x100 pixels which contain a random set of 5 MNIST digits (that can include between 0 to 6 even numbers), randomly positioned within that image (fig \ref{fig:MNISTsample}).
Rejection sampling is used to ensure digits are well separated. Each such image is weakly labeled with the total number of even digits present in that image. The data set is imbalanced and contains significantly more samples showing zero even digits. Thus a naive model which always predicts zero is already able to achieve a better performance than chance level (random-guessing). The model is supposed to learn to count the number of even digits given a weak label in order to solve this task correctly.

For the the Multi-Spike Tempotron the images have to be encoded as spiketrains. We first consider a naive spike-encoding which encodes each individual pixel as $3$s long spiketrain generated by a $\Gamma_5$ process with the rate proportional to the pixel's intensity (grey value). This type of encoding is naive in the sense, that it considers each pixel to be independent and thus does not exploit local spatial correlations of images. Next we consider a more sophisticated spike-encoding frontend, the \textit{Filter-Overlap Correction Algorithm} (FoCal), a model of the fovea centralis \cite{Bhattacharya2010}. This encoding algorithm makes use of spatial correlations in order to reduce the amount of redundant information. This is similar to the convolutional filters embedded in current deep neural networks.
For comparison, we train a conventional ConvNet architecture that has been shown to successfully accomplish this task when trained on $100000$ samples \cite{Fomoro}.  The architecture uses several layers (conv1 - MaxPool - conv2 - conv3 - conv4 - fc - softmax) and includes recently discovered advances like strided and dilated convolutions. To train the ConvNet we use the ADAM \cite{kingma2014adam} optimizer which has been found to be the most effective optimizer for training ConvNets. For the MST model we use our adaptive learning rate method and the originally proposed Momentum method. Since we want to evaluate with regard to computational and sample efficiency all models are trained for 30 epochs on the same training set of 800 images and are evaluated on an independent test set of 800 unseen images.
Perceptually, the counting problem is more similar to a regression problem, since one does not know a-priori the maximum number of desired concepts present in an input. For this reason, we choose \textit{root mean-squared error} (RMSE) of wrongly counted even digits as the evaluation criterion, where a lower value means better performance. This criterion especially penalizes predictions that show a large deviation between the true and the predicted value.
Thus, it is relevant to point out that the ConvNet model is built using prior knowledge about the distribution of the training set. It is constrained to learn a categorical distribution over $[0,6]$, where $6$ is the maximum possible count of even digits in an image. This has two important implications; First, the ConvNet model will be unable to predict the number of even digits in images that include more than $6$ even digits. While for this particular task the data-set is constructed such that this is not possible, in general regression problems the prediction targets are usually not bounded. Second, the error here is constrained as well, by the training bound, to be $6$.
In contrast, the MST model does not have any need for this prior knowledge or constraints. In principle it is capable of solving the general, true regression problem and can also make predictions for images that contain more than 6 even digits. This farther implies, that it solves an even more difficult learning problem. The maximum prediction error in this case is unbounded and it makes the MST more vulnerable, a-priory to high RMSE compared to the ConvNet.
Results are summarized in table \ref{table:resultsMNIST}, the best performing model, MST with adaptive learning rate, is highlighted in bold. We find that generally the MST with adaptive learning rate performs better compared to the Momentum, independent of the choice of a particular spike-encoding front-end. Interestingly the single-neuron MST model is also able to outperform the rate-based ConvNet. In order for the ConvNet to achieve better RMSE ($\sim 0.95$), similar to the MST model, the ConvNet needs to be trained for about 5-10x more epochs than the MST (data not shown). If the model's complexity in terms of free parameters is taken into account (adjusted RMSE), the MST model is even way more computationally efficient.
We find that using FoCal as spike-encoding front-end works much better compared to our naive encoding, which is expected behaviour. It exploits local, spatial correlations, a paradigm known to be more effective compared to pixel-by-pixel consideration. This is in agreement with artificial neural networks where the success of ConvNets over regular, multilayer networks is mostly due to the learned spatial filters by its convolutional layers.
Since we've shown that encoding can impact the model's performance, it is possible that by applying additional efficient encoding algorithms, the performance of the MST model can be improved even farther than the results presented here. We leave the exploration of different types of encodings open for future research.
\begin{table}[t]
      \caption{Results for the Counting MNIST experiment where the model should learn to count the number of even handwritten digits present in a given 100x100 pixel image. We evaluate each model with regard to it's complexity (number of parameters / synapses), and RMSE of wrongly counted digits (lower is better). The ConvNet and MST models have both been trained for 30 epochs on the same training set of 800 samples. Evaluation was done on an independent set of 800 validation samples. For reference we also report performance for naive models which \textit{always-predict zero} and do  \textit{random-guessing}. }
      \label{table:resultsMNIST}
      \centering
      \begin{tabular}{lll}
        \toprule
        \multicolumn{3}{c}{Counting MNIST Results}                   \\
        \cmidrule{1-3}
         Model           &    \#Parameters    &    RMSE \\ 
        \midrule
         ConvNet & $26471$  & $1.49$ \\ 
         3-layer MLP (FoCal enc.) & $960038$  & $1.65$ \\ 
         $MST_{adaptive}$ (naive enc.)    & $10000$ & $2.34$ \\  
         $\mathbf{MST_{adaptive}}$ \textbf{(FoCal enc.)}     & $10000$ & {$\mathbf{1.22}$} \\ 
         $MST_{Momentum}$ (naive enc.)     & $10000$ & $1.87$ \\ 
         $MST_{Momentum}$ (FoCal enc.)     & $10000$ & $1.28$ \\ 
         always-zero     & n/a & $1.65$ \\ 
         random guessing     & n/a & $2.50$ \\ 
        \bottomrule
      \end{tabular}
\end{table}

\section{Discussion}
Rate-based neural networks and algorithms, especially their implementation through deep learning (\cite{lecun2015deep}, \cite{schmidhuber2015deep}), have seen great success in building \textit{intelligent} artificial systems that are able to solve remarkably complex tasks. However, with their increasing rate of success their computational complexity has also been shown to grow exponentially during recent years \cite{OpenAI:compute}. The visual nervous systems of animals are highly efficient in solving complex classification tasks. Therefore it is worth to examine alternative computing models that exploit biological mechanisms of visual information processing. 

In this work we have explored the Multi-Spike Tempotron \cite{Gutig:2016} (MST), a spiking neuron model of the visual cortex (as part of the neocortex), which can be trained by gradient-descent to produce precise number of output spikes in response to a certain stimulus. We first studied and quantified the learning and generalization performance of the model in the general, biological context of task-related spiking activity in the visual cortex. Since the exact spike-train statistic in the visual cortex is still unknown, we specifically studied several, different input statistics that deviate from homogeneous Poisson, which due to its mathematical convenience is the commonly used model of spiking statistics of the visual cortex and has also been used in the original work of the MST model \cite{Gutig:2016}. We showed, that by choosing different and biologically more realistic input statistics, the MST model exhibits large variance with regard to training error and more importantly with regard to generalization on unseen inputs. In order to overcome this issue, we have successfully proposed a modified learning rule that uses adaptive learning rates per synapses and smoothing over past gradients instead of the original Momentum-based learning. We evaluated both methods on data-sets with different input statistics that resemble task-related spiking activity in the visual cortex and under different levels of background noise complexity. We were able to show, that the adaptive learning rate method performs consistently better as compared to Momentum in terms of variability of training error and generalization. The modified learning rule has a regularizing effect and prevents the model from overfitting to the training data, without modifying the model's equation and gradient derivation.

While previous related work of gradient-based learning in spiking network models are mostly concerned with solving classical classification tasks, in this work we applied the single-neuron MST model to solve a regression problem. Specifically, we have used the improved learning rule and applied the MST to the non-biological problem of multiple-instance learning with weakly labeled objects. For this we have used a visual counting task of handwritten even digits from the MNIST data-set. For successful learning, the model needed to solve the binding problem using the weak label and count the number of even digits in each image. Finally we have asked the question, whether a single spiking neuron model is able to compete against more complex, rate-based network models with approximately the same parameter space and training data. For this we have compared the MST against a conventional convolutional neural network of seven layers and assessed the performance using the \textit{root mean-squared error} (RMSE) of wrongly counted digits. We have found, that the improved MST model can outperform the ConvNet in this setting, which needed 5-10x more training epochs to reach the MST performance. While in this work we specifically focused on the computational capabilities of the single-neuron model, the same model and learning rule can also be used to create more complex and layered networks. We leave the study of complex networks of multiple, interconnected Multi-Spike Tempotrons up for farther research.
We conclude that, despite it's simplicity, the single-neuron Multi-Spike Tempotron provides competitive performance not only for biologically relevant inputs like task-related activity in the visual cortex (and neocortex), but also on tasks that are unrelated to Biology. We are confident it can be considered for other classes of machine learning problems that go beyond strict classification tasks. We made our code publicly available to support further research \cite{Rapp:2018}. 
 
\subsubsection*{Acknowledgments}
A significant portion of this work has been conducted during the OIST Computational Neuroscience Course and has been supported by accommodations and travel grants from the Okinawa Institute of Science and Technology (OIST).

\bibliographystyle{unsrt}
\bibliography{bibliography}

\end{document}